\documentclass[3p,times,procedia]{elsarticle}
\usepackage{ecrc}
\usepackage[bookmarks=false]{hyperref}
    \hypersetup{colorlinks,
      linkcolor=blue,
      citecolor=blue,
      urlcolor=blue}
\usepackage{graphicx}
\usepackage{subfig}
\usepackage{amssymb,amsmath}
\usepackage{ecrc}
\usepackage[capitalise]{cleveref}
\usepackage{multirow}
\firstpage{1}

\journalname{Procedia Computer Science}

\runauth{A. Bubnova et al.}
\runtitle{MIxBN}

\jid{procs}

\usepackage{amssymb}
\usepackage[figuresright]{rotating}
\begin{document}

\begin{frontmatter}



%\dochead{17th International Conference on Dynamical Processes in Excited States of Solids}

\title{MIxBN: library for learning Bayesian networks from mixed data}


\author[a]{Anna~V.~Bubnova\corref{cor1}} 
\author[a]{Irina~Deeva}
\author[a]{Anna~V.~Kalyuzhnaya}

\address[a]{ITMO University, Saint-Petersburg, Russia}

\begin{abstract}
This paper describes a new library for learning Bayesian networks from data containing discrete and continuous variables (mixed data). In addition to the classical learning methods on discretized data, this library proposes its algorithm that allows structural learning and parameters learning from mixed data without discretization since data discretization leads to information loss. This algorithm based on mixed MI score function for structural learning, and also linear regression and Gaussian distribution approximation for parameters learning. The library also offers two algorithms for enumerating graph structures - the greedy Hill-Climbing algorithm and the evolutionary algorithm. Thus the key capabilities of the proposed library are as follows: (1) structural and parameters learning of a Bayesian network on discretized data, (2) structural and parameters learning of a Bayesian network on mixed data using the MI mixed score function and Gaussian approximation, (3) launching learning algorithms on one of two algorithms for enumerating graph structures - Hill-Climbing and the evolutionary algorithm. Since the need for mixed data representation comes from practical necessity, the advantages of our implementations are evaluated in the context of solving approximation and gap recovery problems on synthetic data and real datasets.
\end{abstract}

\cortext[cor1]{- corresponding author (avbubnova@itmo.ru)}
\end{frontmatter}

%\correspondingauthor[*]{Corresponding author. Tel.: +0-000-000-0000 ; fax: +0-000-000-0000.}

% \linenumbers


%\enlargethispage{-7mm}
\section{Introduction}
 Often real data contain variables of different types, e.g. discrete and continuous. It is essential to use the right tools for dealing with such mixed data if you need to train Bayesian networks. Hybrid Bayesian model \cite{bottcher2001learning} is an excellent tool to analyze the complex dependence structure without involving experts and solving practical problems, such as gap recovery, prediction, and anomaly search. However, their applicability is limited by the quality of their learning. Although there are many flexible and well-proven approaches to parameter learning \cite{bottcher2001learning}, structure learning is not easy. 
 
A typical situation is when the structure learning stage is much more labour- and resource-intensive than the parameter learning stage because the number of possible DAGs grows superexponentially depending on the number of nodes \cite{robinson1973counting}. There are many different approaches \cite{scutari2019learns} to solve this problem, which can mainly be divided into those that rely on independence tests and those that use some score functions. In both cases, however, the most common algorithms are those that deal either only with continuous variables \cite{baba2004partial, zhang2012kernel, geiger1994learning} or only with discrete variables \cite{agresti2002categorical, cooper1992bayesian}. The first group is not suitable for our situation and rather refers to Gaussian Bayesian networks, while the second group requires discretization of continuous variables. As a result, the problem becomes more complicated since we need to choose the right number of bins and method, and some information is lost. Estimation also does not consider how accurately we can capture the parameters of the distributions in the next stage of learning.

Fortunately, algorithms using functions and statistical methods suitable for mixed data have recently begun to appear. For example, based on new independence tests \cite{tsagris2018constraint} or modified score functions: log-likelihood and BIC \cite{andrews2018scoring}.

Considering the described situation, we developed our own library MIxBN for learning Bayesian networks on mixed data without discretization. A key feature of the library is an internally developed algorithm for mixed BN learning that produces structured learning based on a new modification of mutual information (MI) score function and entropy for mixed data. Also this algorithm takes mixed data and takes into account the specifics of mixed parameters learning, which relies on the approximation of continuous variables by Gaussian distributions. Also, such a modification of MI can be used to compute log-likelihood, AIC, and BIC score functions since they are all determined by MI and entropy. In the future, this may form the basis of a group of structural learning algorithms, which may differ in efficiency from the analogues proposed in the paper \cite{andrews2018scoring}.

\begin{figure}[ht!]\vspace*{4pt} 
\centerline{\includegraphics[width=0.7\textwidth]{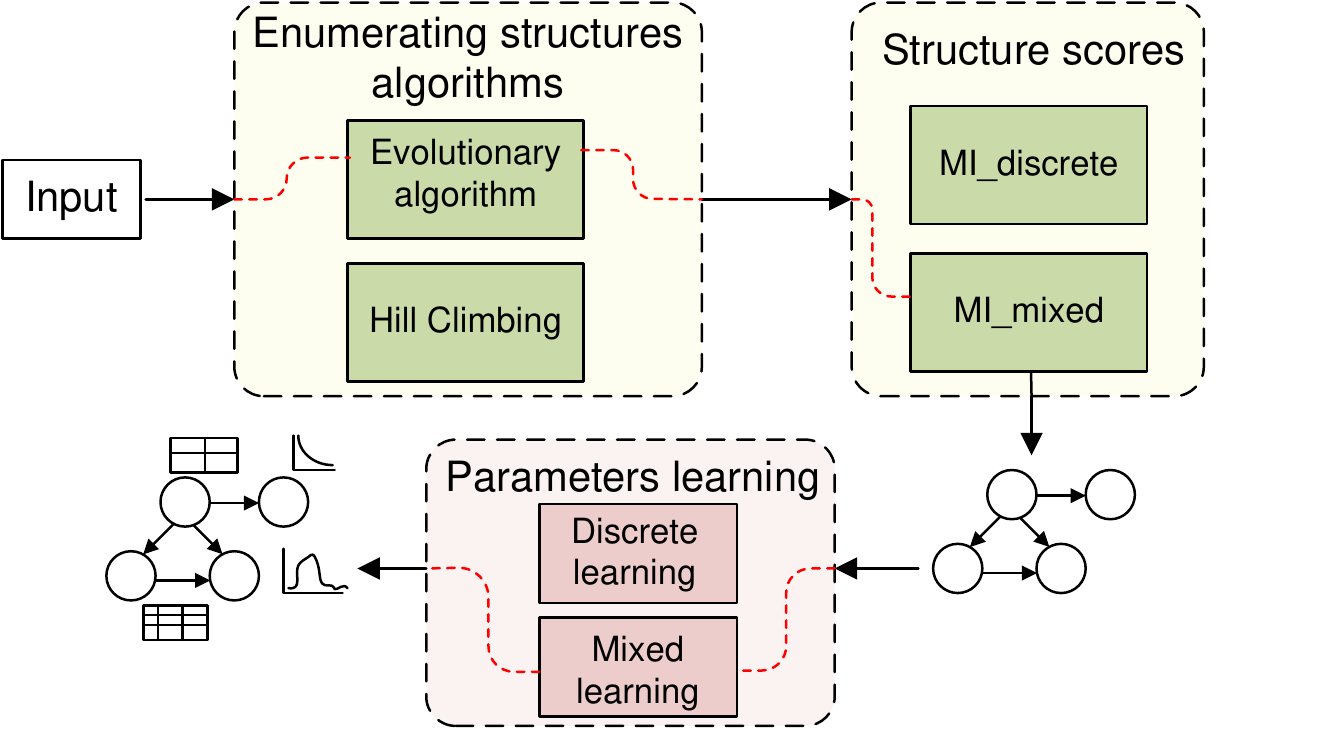}}
\caption{MIxBN library schema. The red dashed line shows an example of the choice of algorithms at each step of BN learning.} \label{fig:pipeline}

\end{figure}

Also, our library allows you to work with two algorithms for enumerating graph structures: Hill-Climbing and evolutionary algorithm. They describe only the walking procedure in the space of possible DAGs and can take both discrete and mixed MI variants. This will allow us to compare the advantages and disadvantages of these two approaches. Separately, it is worth considering the parameter learning procedure, which, depending on the type of distributions in the approximation, can also be discrete or continuous. A scheme of the proposed library is shown in \cref{fig:pipeline}. 

\begin{nomenclature}
\begin{deflist}[WMO]
\defitem{BN}\defterm{Bayesian network}
\defitem{HC}\defterm{Hill-Climbing}
\defitem{DAG}\defterm{Directed acyclic graph}
\defitem{MI}\defterm{Mutual information}
\defitem{AIC}\defterm{Akaike information criterion}
\defitem{BIC}\defterm{Bayesian information criterion}
\end{deflist}
\end{nomenclature}

\section{Related Work}\label{Related}
This section will present brief information and references to works on structural learning algorithms, in part using mixed score functions.

The closest to our work considers a mixed BIC function and its modifications \cite{andrews2018scoring}. In this work, fGES \cite{ramsey2017million} is taken as the enumerating DAG structures algorithm. However, it imposes certain restrictions on the list of available functions, as they must have local consistency \cite{chickering2002optimal}. For example, MI does not have this property, and there is a certain problem with considering it under the same conditions. Fortunately, Hill-Climbing and evolutionary algorithms do not require it. 

 Many approaches with discrete score functions for structure learning are compared in \cite{scutari2019learns}. Unfortunately, mixed functions are not represented there. This is unexpected because the bnlearn package the author is working on contains function variants suitable for working with hybrid Bayesian networks \cite{scutari2020package}. For example, there are the conditional linear Gaussian log-likelihood and corresponding AIC, BIC functions.

We should also mention the deal package for structure learning on mixed data using a posteriori score \cite{bottcher2003deal}. However, this function belongs to the Dirichlet score class rather than to MI and the similar ones.

Not all existing packages for learning Bayesian networks are capable of implementing structural and parameters learning on mixed data, as well as offering various approaches to enumerating graph structures such as evolutionary algorithms, for example \cite{scanagatta2019survey}. \cref{package} shows a comparison of the most popular packages for working with Bayesian networks in terms of their ability to work with mixed data and different algorithms for enumerating graph structures.

\begin{table}[ht!]
\caption{Comparison of the most popular packages for Bayesian networks in terms of the ability to work with mixed data.}
\label{package}
\begin{center}
\begin{tabular}{|c|c|c|c|c|}
\hline
Package     & \begin{tabular}[c]{@{}c@{}}Structure learning \\ on mixed data\end{tabular} & \begin{tabular}[c]{@{}c@{}}Parameters learning \\ on mixed data\end{tabular} & \begin{tabular}[c]{@{}c@{}}Using evolutionary \\ algorithms\end{tabular} & \begin{tabular}[c]{@{}c@{}}Using MI for \\ mixed data\end{tabular} \\ \hline
bnlearn     & +                                                                           & +                                                                            & -                                                                        & -                                                                  \\ \hline
libpgm      & -                                                                           & +                                                                            & -                                                                        & -                                                                  \\ \hline
pomegranate & -                                                                           & -                                                                            & -                                                                        & -                                                                  \\ \hline
DEAL        & +                                                                           & +                                                                            & -                                                                        & -                                                                  \\ \hline
pgmpy       & -                                                                           & -                                                                            & -                                                                        & -                                                                  \\ \hline
\end{tabular}
\end{center}
\end{table}

\section{Backgrounds}
In general, Bayesian networks are a way to represent a factorization of probability distribution $P(X_1,\dots,X_p)$ using a Directed Acyclic Graph (DAG) $G = (V, E)$, where $V=\{X_j\}_{j=1}^p$. By factorization in this case we mean the decomposition of joint probabilities into the product of conditional probabilities. The conditions are the sets of parents $\Pi_X=\{X_k:X_k\neq X, (X_k,X)\in E\}$ generated by the structure of graph $G$:
\begin{equation}
    P(X_1,\dots,X_p )=\prod_{j=1}^p P(X_j |\Pi_{X_j}).
\end{equation}
In this paper, we focus on the group of algorithms for finding $G$ from the available data $D$, which best describe the factorization of the observed joint distribution relying on score functions. From a formal point of view, such algorithms walk in DAG space and look for the maximum or minimum sum of some local score function $s$:
\begin{equation}
    S(G, D)=\sum_{j=1}^p s(X_j|\Pi_{X_j}) \rightarrow \max(\min).
\end{equation}

In this context, our goal is to implement and compare the quality and speed of algorithms that learn a structure without discretization based on a mixed score function MI and functions involving pre-discretization such as classical MI. 

\section{Algorithms and methods}
\subsection{Structure learing}

\subsubsection{HC and evolutionary algorithms}\label{HC and EVO}

In structured learning of a Bayesian network, strategies to enumerate possible structures play a key role. The most straightforward approach is a complete enumeration of all possible structures; however, since the number of these structures grows super-exponentially depending on the number of nodes \cite{robinson1973counting}, such an enumeration is not possible. This study looked at two main strategies - greedy heuristics (Hill-Climbing algorithm) and generic algorithms. 

Hill-Climbing algorithm has long been used to learn structures \cite{chickering2002optimal, scanagatta2019survey}. The essence of this algorithm is that at each iteration, we check the possible action with the edge (add, delete, reverse) and remember the structure that gives the greatest increase in the score function. The main disadvantage of this strategy is that we risk finding only a local optimum since we maximize the score function only locally. Evolutionary algorithms are also used for structured learning \cite{larranaga2013review}. For this study, we used an algorithm based on the evolutionary algorithms of the FEDOT framework \cite{nikitin2020structural}. The evolutionary algorithm is based on the following steps:
\begin{enumerate}
\item Generation of a random population of graphs;
\item Application of mutation and crossover operators;
\item Selection of a part of individuals from the population with subsequent repetition of steps;
\item Stop when the stop condition is reached.

\end{enumerate}

Since the FEDOT framework is an automatic machine learning framework, models are found at the nodes of the obtained graphs. This imposes certain features on the evolutionary algorithm for obtaining optimal graphs. For our problem, restrictions were imposed on the FEDOT evolutionary algorithm - nodes cannot be duplicated, continuous nodes cannot be parents of discrete nodes.

\subsubsection{Mutual information calculation function}\label{Mixed MI}
A special feature of this work is MI for mixed data and other functions based on it in the structural learning phase. For sets of discrete values, it is calculated absolutely the same as the classical version. However, it relies on an approximation of continuous with Gaussian distributions and differential entropy for Gaussian for fully continuous and mixed. Below we present the formulas used to calculate the mixed MI.

The basic formula for calculating MI using entropy remains unchanged:
\begin{equation}\label{MI}
MI(X,Y)=H(X)-H(X|Y).
\end{equation}
Differential entropy of a multivariate normal distribution $N(\mu, \Sigma)$ is calculated as follows:
\begin{equation}
H(N(\mu, \Sigma)) = \frac{1}{2}\ln{|2\pi e \Sigma|}.
\end{equation}
Since $MI(X)=H(X)$ to calculate MI for a set of continuous variables, we can estimate the covariance matrix $\Sigma$ for the corresponding multivariate value from the sample and calculate entropy using the formula.

In the mixed case, however, the total set is split into continuous multivariate value $X$ and multivariate categorical value $Y$ and the following formula is used:
\begin{equation}
H(X|Y)=\sum_j P(Y=y_j)H(X|Y=y_j).
\end{equation}
$P(Y=y_j)$ and the covariance matrix $\Sigma_j$ for $H(X|Y=y_j)$ are estimated from the subsample where $Y=y_j$. Then $MI(X,Y$) can be calculated with \eqref{MI}:
\begin{equation}\label{H_cond}
MI(X,Y)=H(X)-H(X|Y)=\frac{1}{2}\ln{|2\pi e \Sigma|}-\frac{1}{2}\sum_j P(Y=y_j)  \ln{|2\pi e \Sigma_j|}.
\end{equation}
In the context of structural learning of Bayesian networks, the function is given by $(X,\Pi_X)$ for node $X$. The local structure does not matter, since it is taken into account at the stage of estimation of probabilities and covariance matrices. The score for the whole network is found by summing the value for each $(X,\Pi_X)$. 
Note that the local log-likelyhood for the set $(X,\Pi_X)$ can be easily determined through the mixed MI:
\begin{equation}
LL(X,\Pi_X )=MI(X, \Pi_X )-H(X).
\end{equation}
And through it BIC and AIC are determined, while the penalty can be calculated according to this work \cite{andrews2018scoring}.

There is one special situation worth paying attention to when calculating \eqref{H_cond}. In the case where only one observation is given at $Y=y_j$ the entropy will be infinitely large, since there is no information about the mean or variance. This situation cannot be ignored because monotonicity will be lost for MI. However, in the case of BIC and AIC it is necessary to do this, or to replace such entropy with one estimated from the full sample. Otherwise, the penalty for system complexity will be disproportionate to the rest of the score.

\subsection{Parameters learning}\label{Param}
For parameters learning of Bayesian networks, we used the principle of likelihood maximization. With the available dataset $D$, it is necessary to select an estimate for $\theta$ parameter that satisfies the condition:

\begin{equation}
L({\stackrel{\wedge}{\theta}} : D) = \max_{\theta \in \Theta}L(\theta : D)
\end{equation}

For a multinominal distribution, such an estimate will be the probability of each value of X corresponding to its frequency in the training data. A continuous distribution is usually approximated by a Gaussian distribution, and in this case, the parameter estimates will be estimates of the mathematical expectation and standard deviation. In the case of estimating parameters at the nodes of a Bayesian network, the very structure of a Bayesian network allows us to split our distribution into a product of conditional distributions (CPD). Suppose we have a dataset $D$ with observations $X_1 ... X_n$, also let $G$ be a Bayesian network structure. Suppose that the parameters $\theta_{X_i | Pa_{X_i}}$ do not intersect with the parameters $\theta_{X_j |Pa_{X_j}}$ for all $j \neq i$. Let $\stackrel{\wedge}{\theta}_{X_i | Pa_{X_i}}$ be those parameters that maximize the local likelihood $L_i (\theta_{X_i |Pa_{X_i}}:D)$. Then $\stackrel{\wedge}{\theta} = (\stackrel{\wedge}{\theta}_{X_1 | Pa_{X_1}} ... \stackrel{\wedge}{\theta}_{X_n | Pa_{X_n}})$ maximizes the global likelihood $ L (\theta : D)$.

\begin{figure}[ht!]\vspace*{4pt}
\centerline{\includegraphics[width=0.7\textwidth]{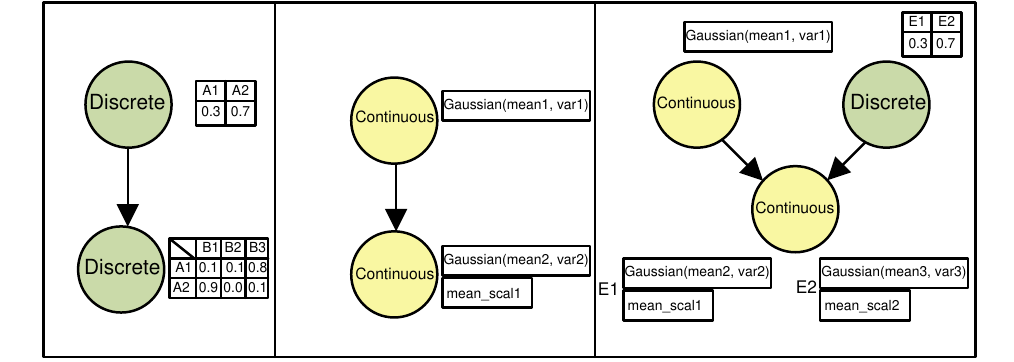}}
\caption{Parameters learning scheme. $Mean scal$ is a linear regression coefficient that takes the parent continuous node values as arguments and the child continuous node values as the target variable. $Mean$ of children nodes is an intercept of the linear regression and $var$ is an error of residuals.}
\label{param_scheme}
\end{figure}

Since we are dealing with mixed data, the distributions themselves will change depending on the type of the node and its parents. If we are dealing with a discrete node and its discrete parents, we learn a table of conditional probabilities. If the node is continuous and has continuous parents, we learn a Gaussian distribution. If the node is continuous and has both discrete and continuous parents, then we learn a conditional Gaussian distribution for all combinations of discrete parents. In the case of a continuous node, we assume that the child node is a linear combination of continuous parents, so the distribution parameters of the child node are also linear combinations of the parent node. In this case, with continuous parents, we train linear regression on continuous parents and then use the coefficients of this regression to obtain the parameters of the child node.\cref{param_scheme} shows a diagram of the distributions that we have.

\section{Experiments and Results}
\subsection{Datasets Description}
The methods will be compared on five mixed datasets: three synthetic and two real datasets. Datasets HEALTHCARE, SANGIOVESE and MEHRA represent the synthetic data sampled from $bnlearn$ networks \cite{scutari2020package}. First, the advantage of such data is that they are accurately described by conditional Gaussian Bayesian networks, which corresponds to the assumptions of structural and parameters learning algorithms. And second, the network for them is known with exact equivalence, which allows us to compare structures. 

However, it is clear that we are interested in BN mainly in the practical sense, so we need to check the efficiency for real cases. The first of these datasets represents various petrophysical characteristics of oil and gas reservoirs. The second one contains information aggregated from users' social media profiles. Both datasets contain anomalies and noise, and for some of the continuous features, we cannot talk about gaussianity. Also, the real networks are unknown, so we cannot use comparisons at this level. Nevertheless, it is possible to assess the extent to which one or the other method can solve practical problems, such as the recovery of missing values. More information about datasets is presented in the \cref{table_datasets'_info}.

\begin{table}[!ht]
\caption{Number of nodes and observations for each dataset.}
\label{table_datasets'_info}
\begin{center}
\begin{tabular}{|c|c|c|c|c|c|c|}
\hline
Parameters & HEALTHCARE & SANGIOVESE & MEHRA & Reservoirs & Social \\  
\hline
Total nodes & 7 & 15 & 24 & 10 & 8\\
\hline
Discrete  nodes & 3 & 1 & 8 & 5 & 4\\
\hline
Continuous nodes & 4 & 14 & 16 & 5 & 4\\
\hline
Size & 3000 & 3000 & 3000 & 514 & 3789 \\
\hline
\end{tabular}
\end {center}
\end{table}

Since we are comparing mixed and discrete methods, it should be mentioned that the second methods use pre-discretization of the data. In our work, each continuous value was divided into five bins with approximately equal frequency, and then discrete methods worked with the labels of these bins as categorical data. Clearly, this may not be the most successful discretization, but the difficulty of choosing appropriate parameters and discretization methods is one of the disadvantages of discrete methods.
    
\subsection{Comparison of Bayesian network learning algorithms} \label{experiments} 
\subsubsection{Comparison of learning approaches on discretized and mixed data}
\begin{figure}[!ht]
\centering
    \subfloat[]{\label{geo}\includegraphics[width=0.33\textwidth]{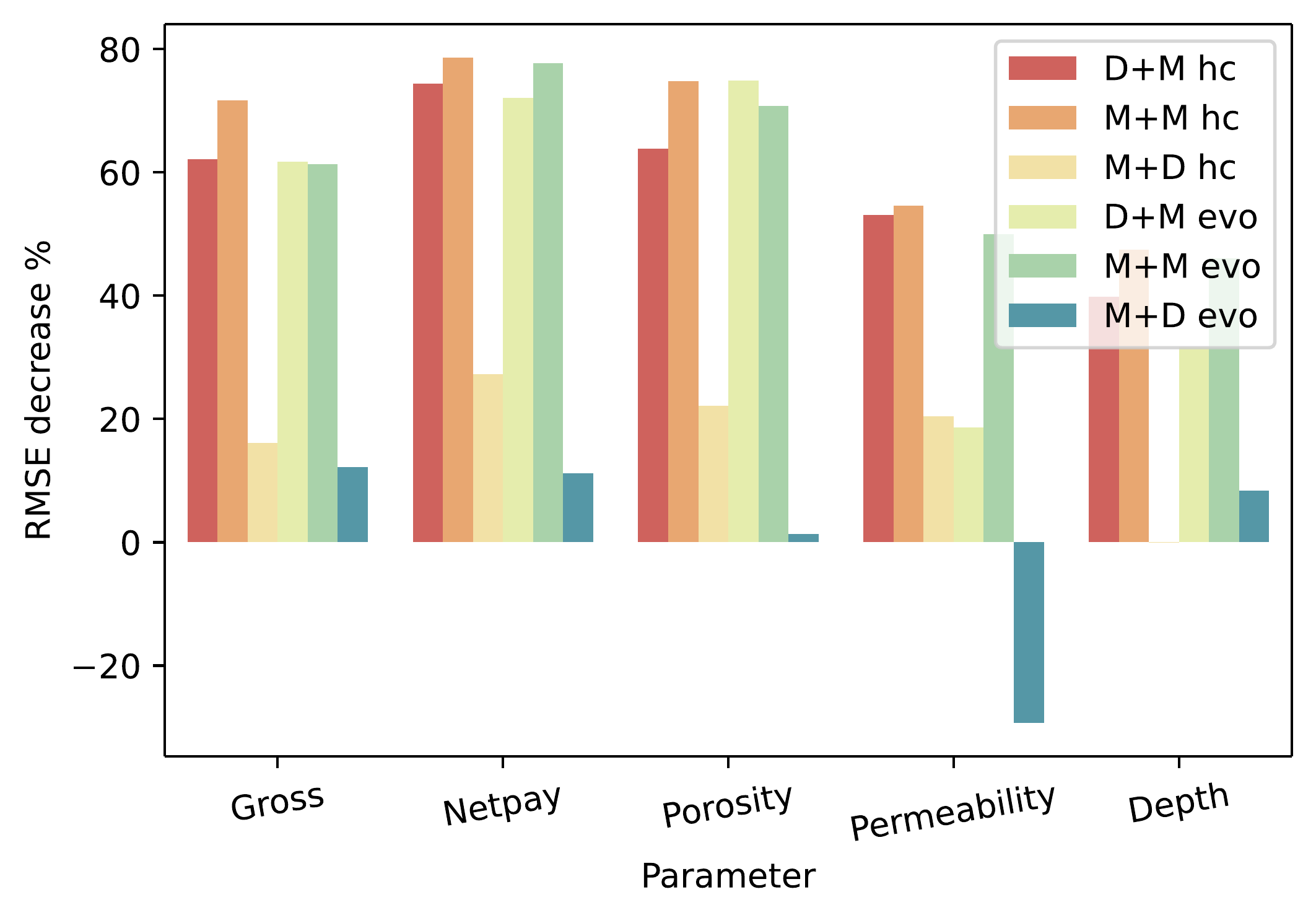}} 
    \subfloat[]{\label{socio}\includegraphics[width=0.33\textwidth]{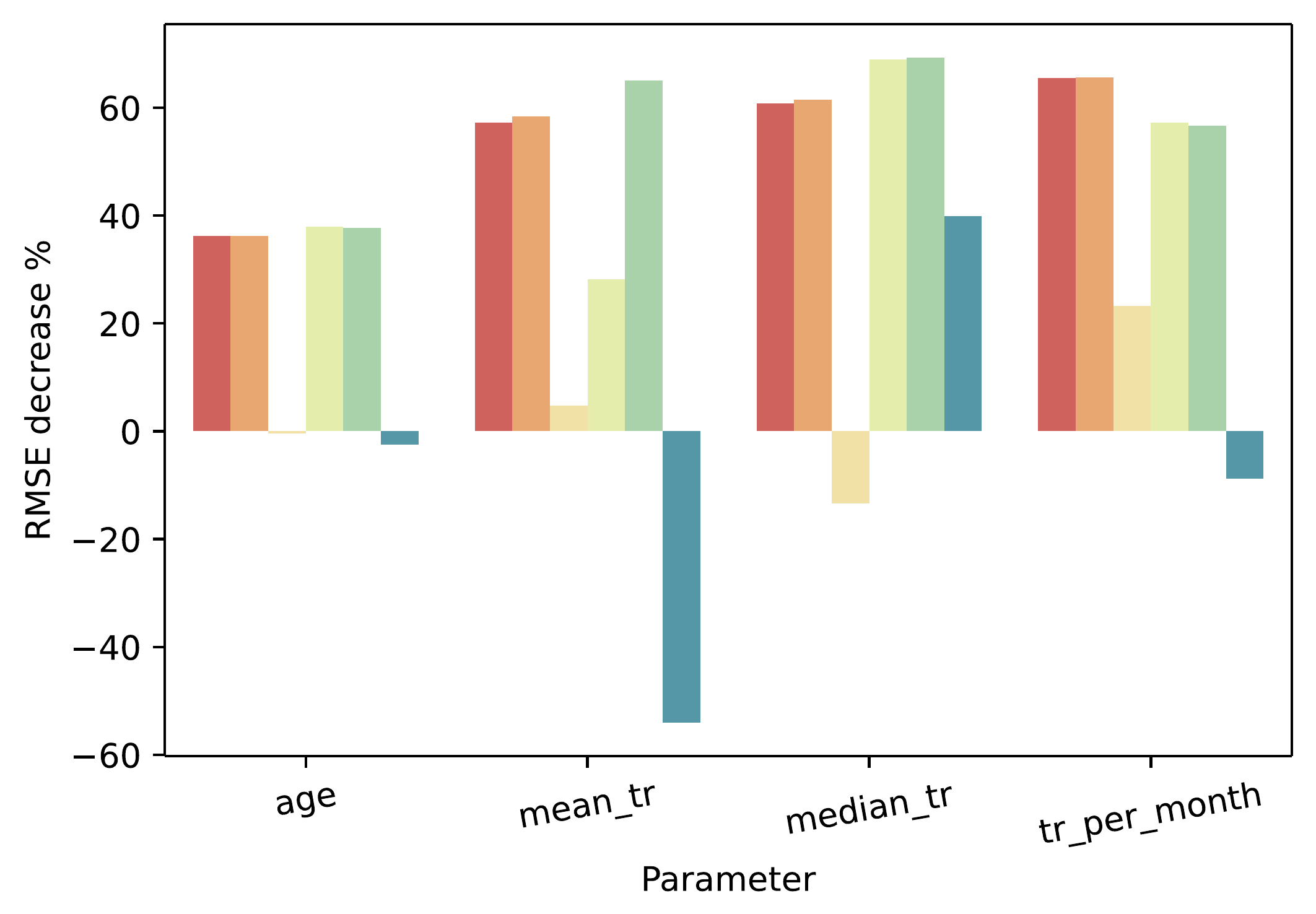}}
    \subfloat[]{\label{health}\includegraphics[width=0.33\textwidth]{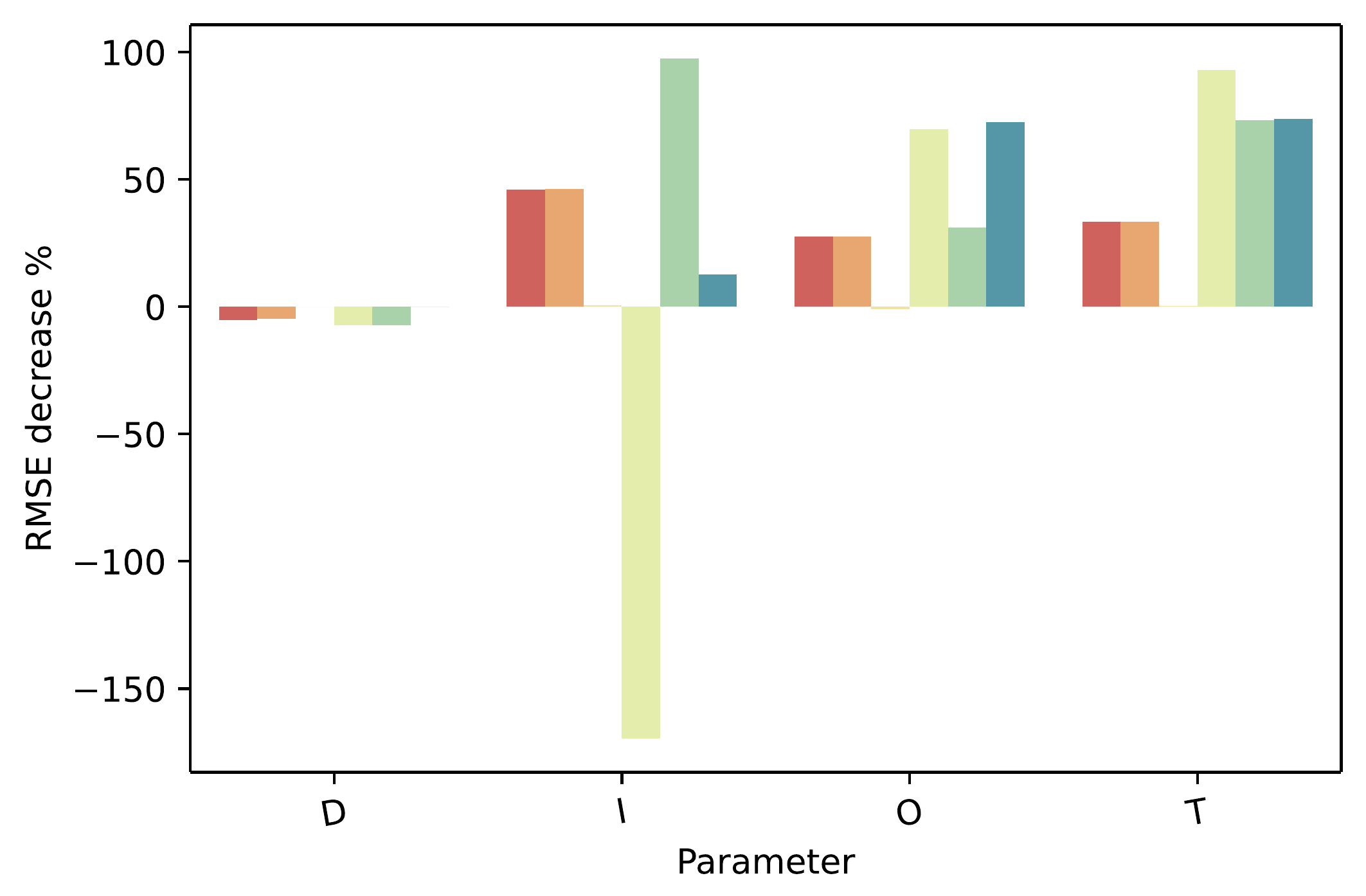}}\\
    \subfloat[]{\label{mehra}\includegraphics[width=0.33\textwidth]{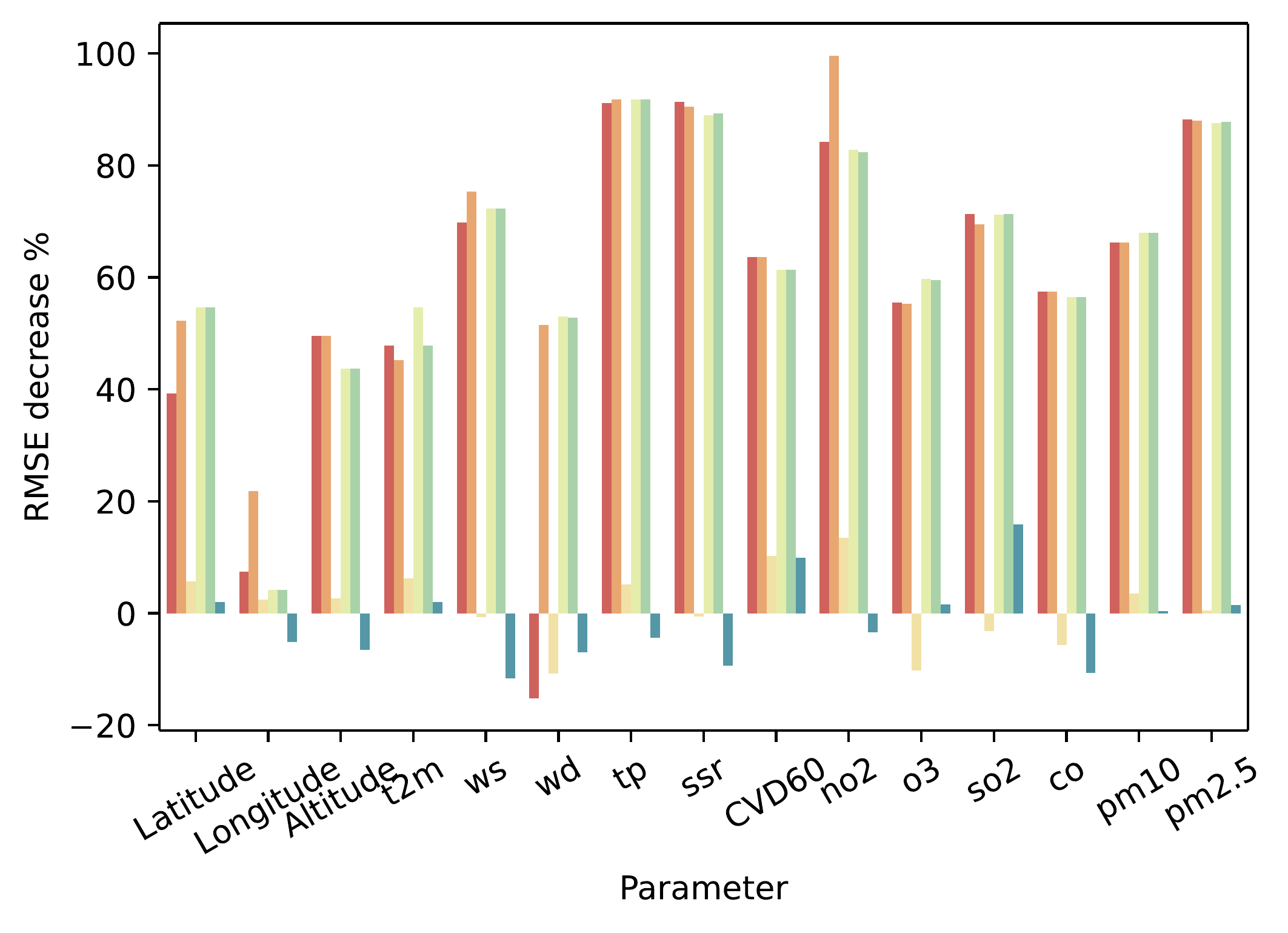}} 
    \subfloat[]{\label{sangiovese}\includegraphics[width=0.33\textwidth]{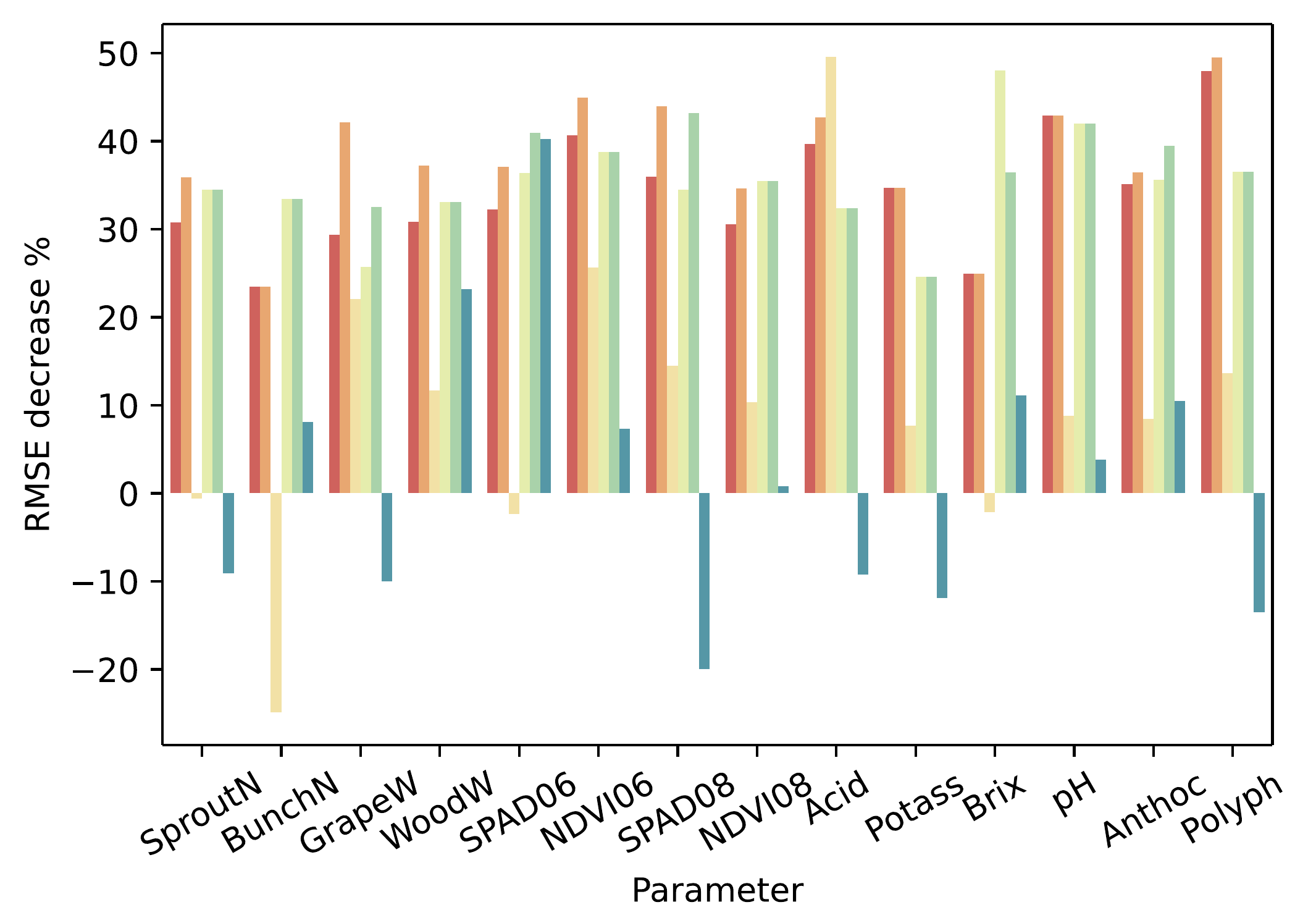}}
    \caption{The percentage of error reduction in comparison with the baseline approach. The blue lines are an evolutionary algorithm, the red lines are Hill-Climbing. In the legend, the first D / M means learning the structure on discretized or mixed data and the second D / M means learning parameters on discretized or mixed data.}
    \label{fig:rmse}
\end{figure}

The proposed approaches to learning a Bayesian network were compared with each other in terms of the restoration accuracy at the nodes of the network. The datasets were divided into training and test samples (in a ratio of 90 to 10 per cent). Then the Bayesian network was learned using the selected algorithm on the training set. In the test dataset, the value in the parameter was deleted and then restored using sampling from BN node. Then the average recovery accuracy was calculated over the entire test dataset. As a baseline approach, an approach was chosen to learn the structure and parameters on discretized data (D+D). Since the proposed algorithms relate specifically to continuous data, we compared the increase in accuracy for each algorithm for continuous parameters. \cref{fig:rmse} shows for each dataset how many per cent the error is reduced compared to the baseline approach. It can be seen that the approach, where both the structure and the parameters are learned on mixed data (M+M), always wins before the baseline solution, and this is typical for both the Hill-Climbing algorithm and the evolutionary one. Also, negative percentages can be observed on the plots - this means that a certain method for a certain parameter loses to the baseline solution. However, it can be seen that this happens only for those approaches where either the structure is learned on discretized data or parameters, which is understandable since information is lost during discretization. However, we repeat that the approach with mixed learning of both parameters and structure (M+M) always wins.
\subsubsection{Comparison of Hill-Climbing and evolutionary algorithm}
\begin{figure}[!ht]
\centering
    \subfloat[]{\label{geo}\includegraphics[width=0.33\textwidth]{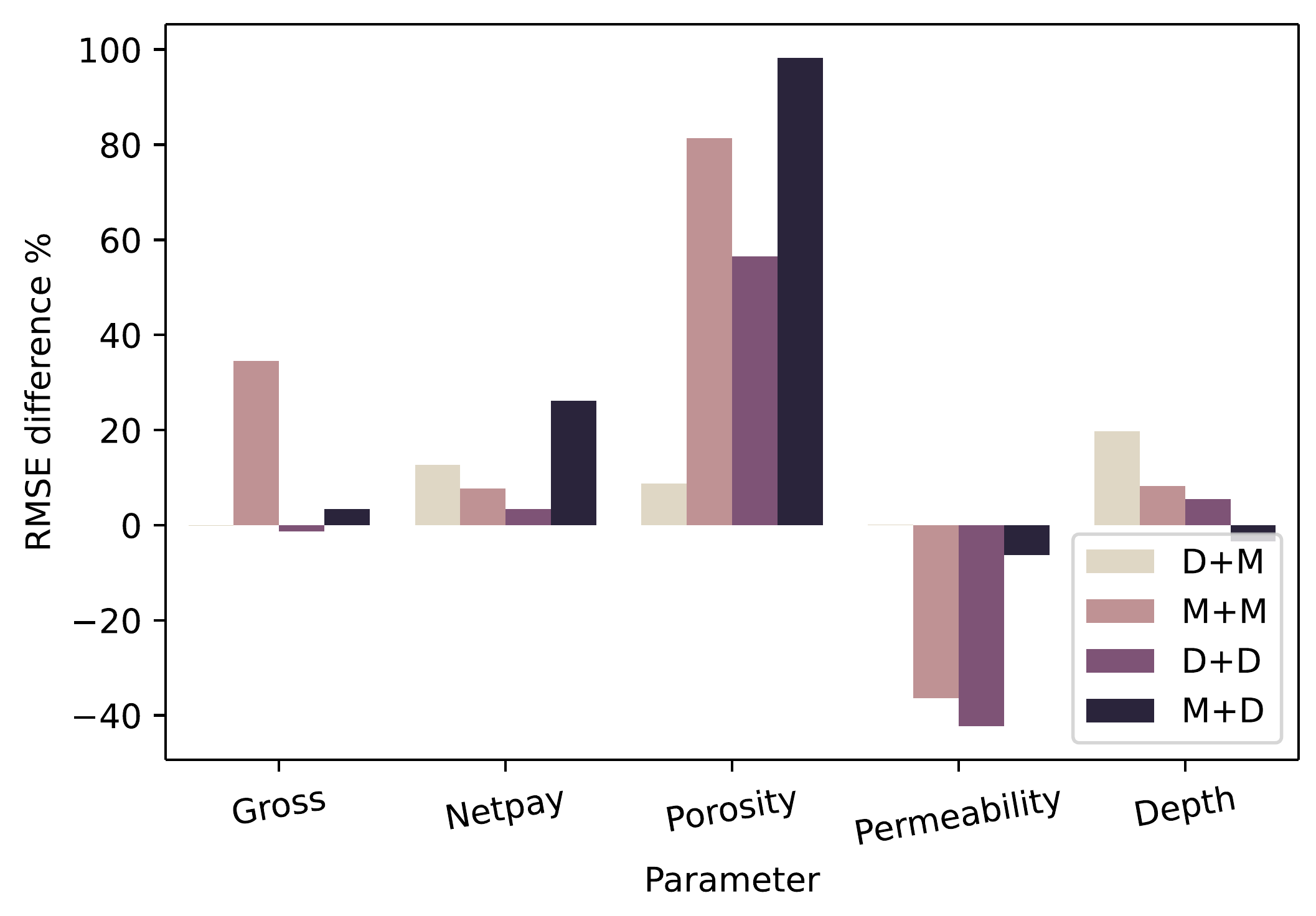}} 
    \subfloat[]{\label{socio}\includegraphics[width=0.33\textwidth]{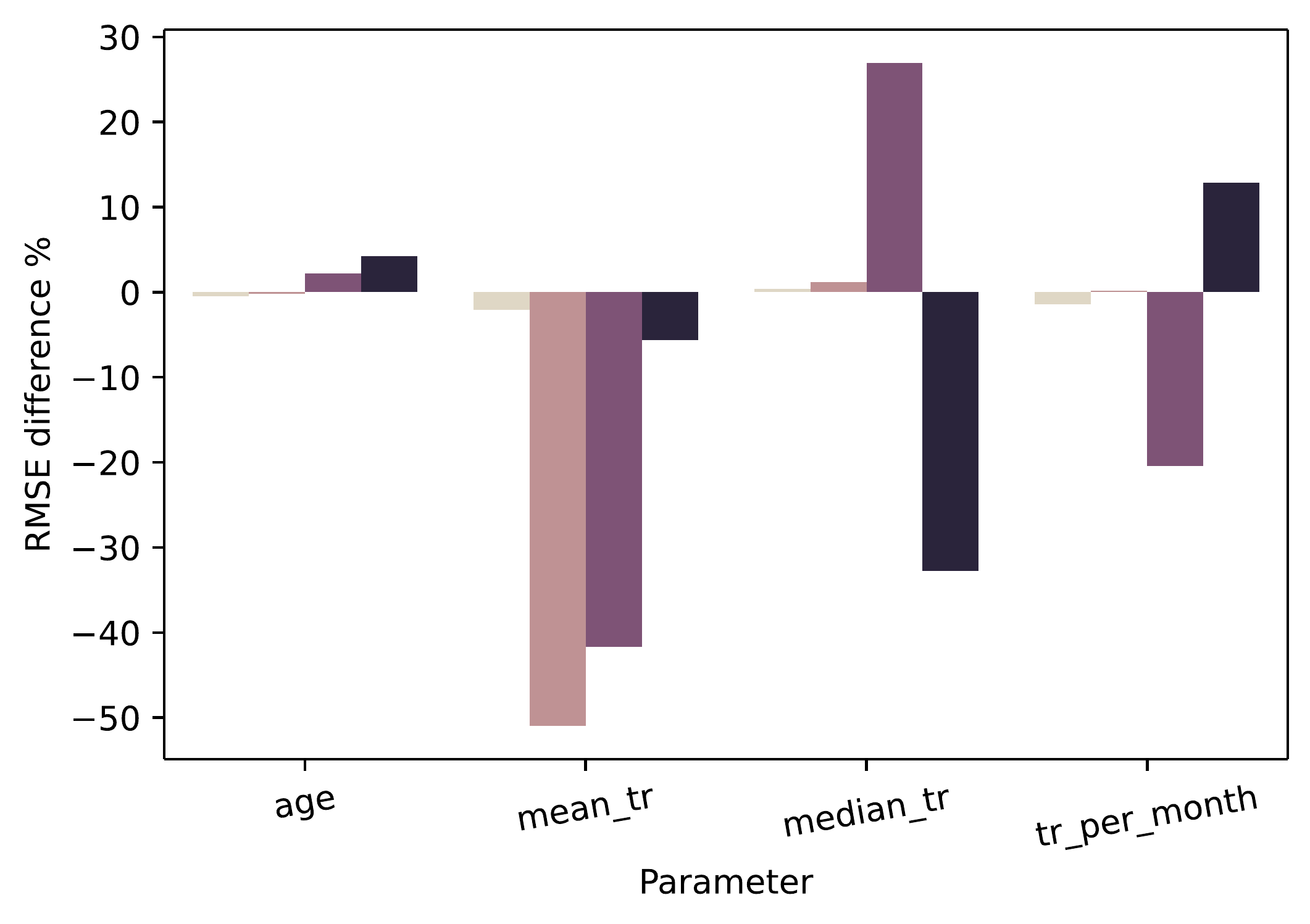}}
    \subfloat[]{\label{health}\includegraphics[width=0.33\textwidth]{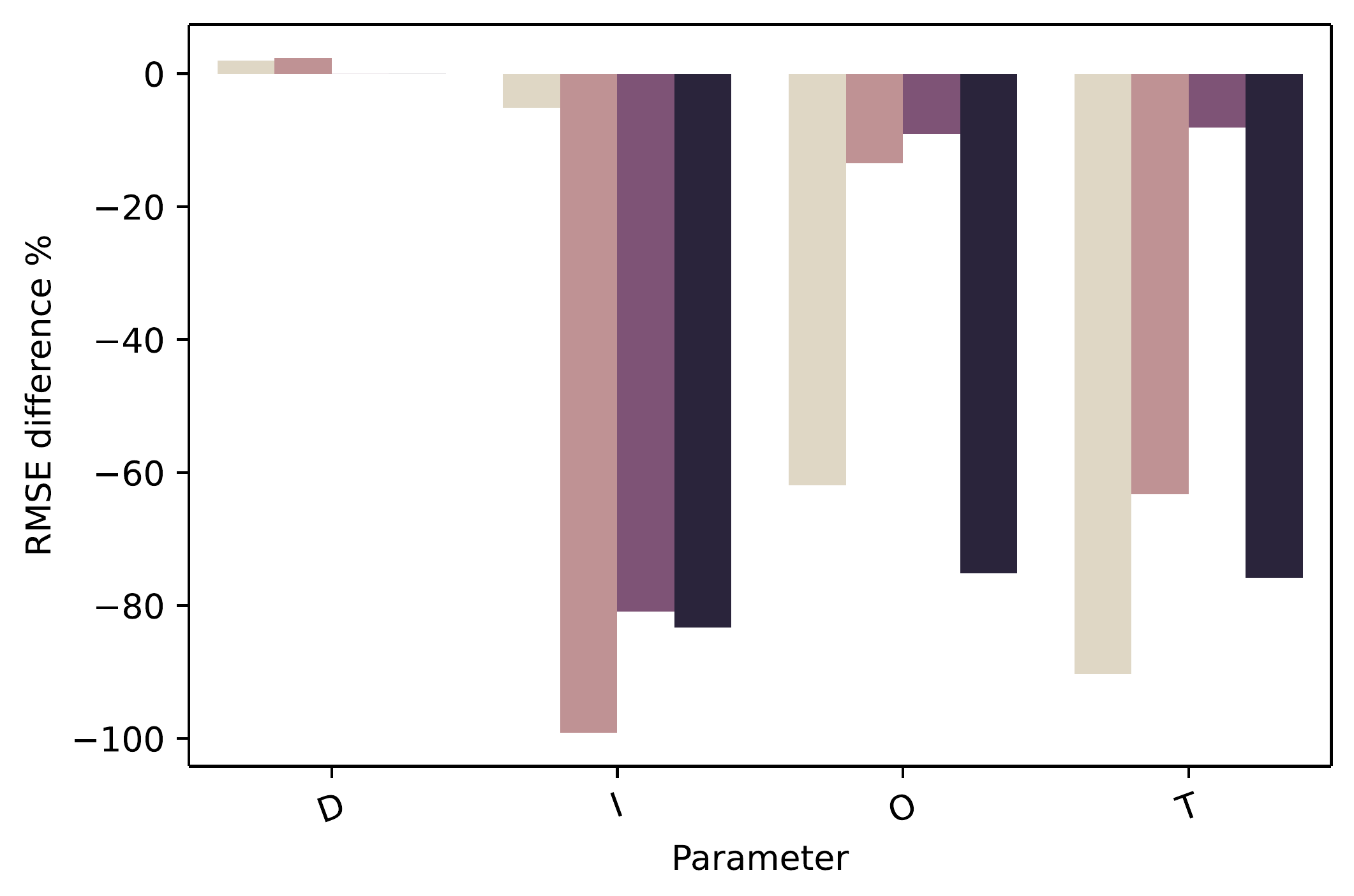}}\\
    \subfloat[]{\label{mehra}\includegraphics[width=0.33\textwidth]{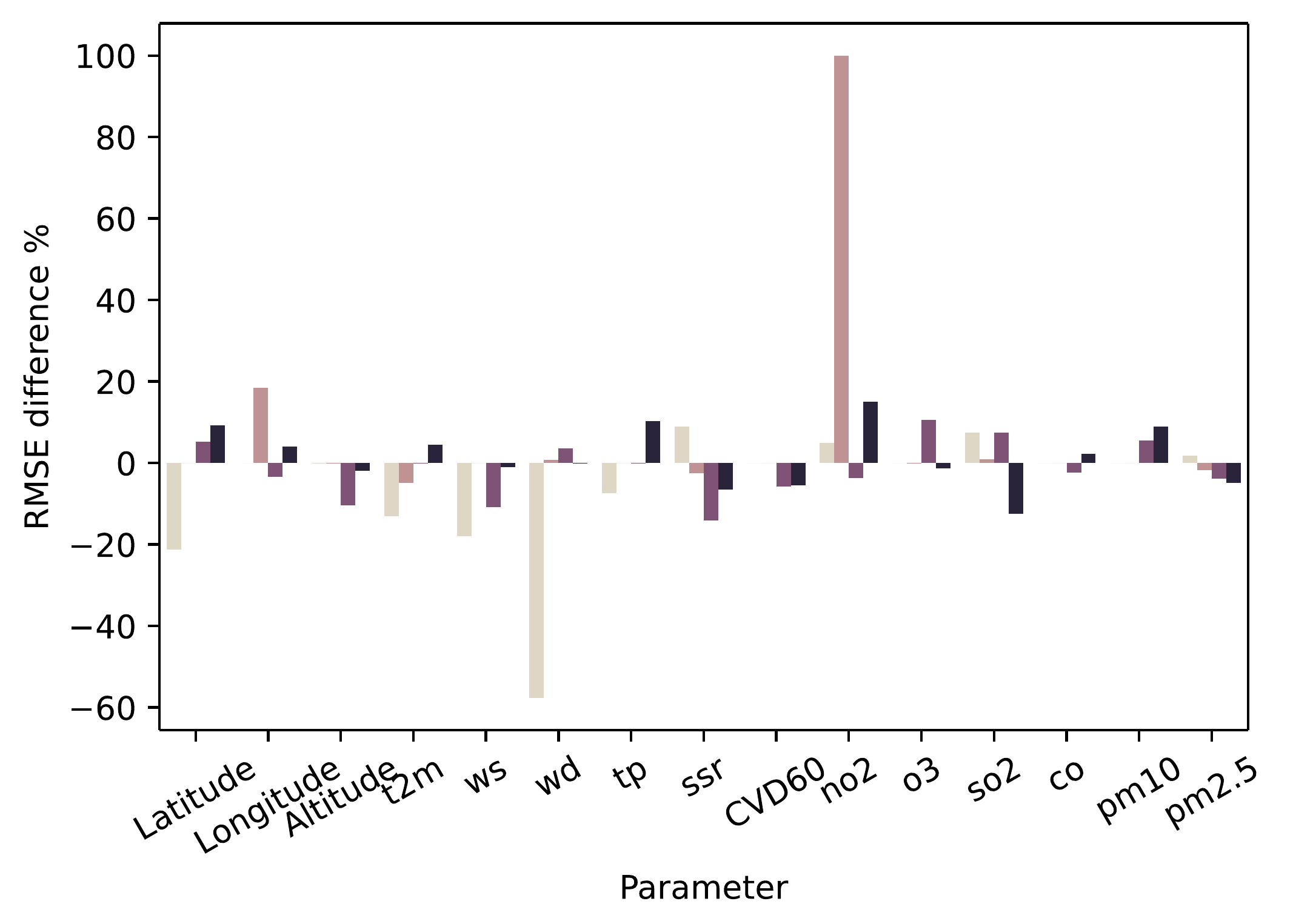}} 
    \subfloat[]{\label{sangiovese}\includegraphics[width=0.33\textwidth]{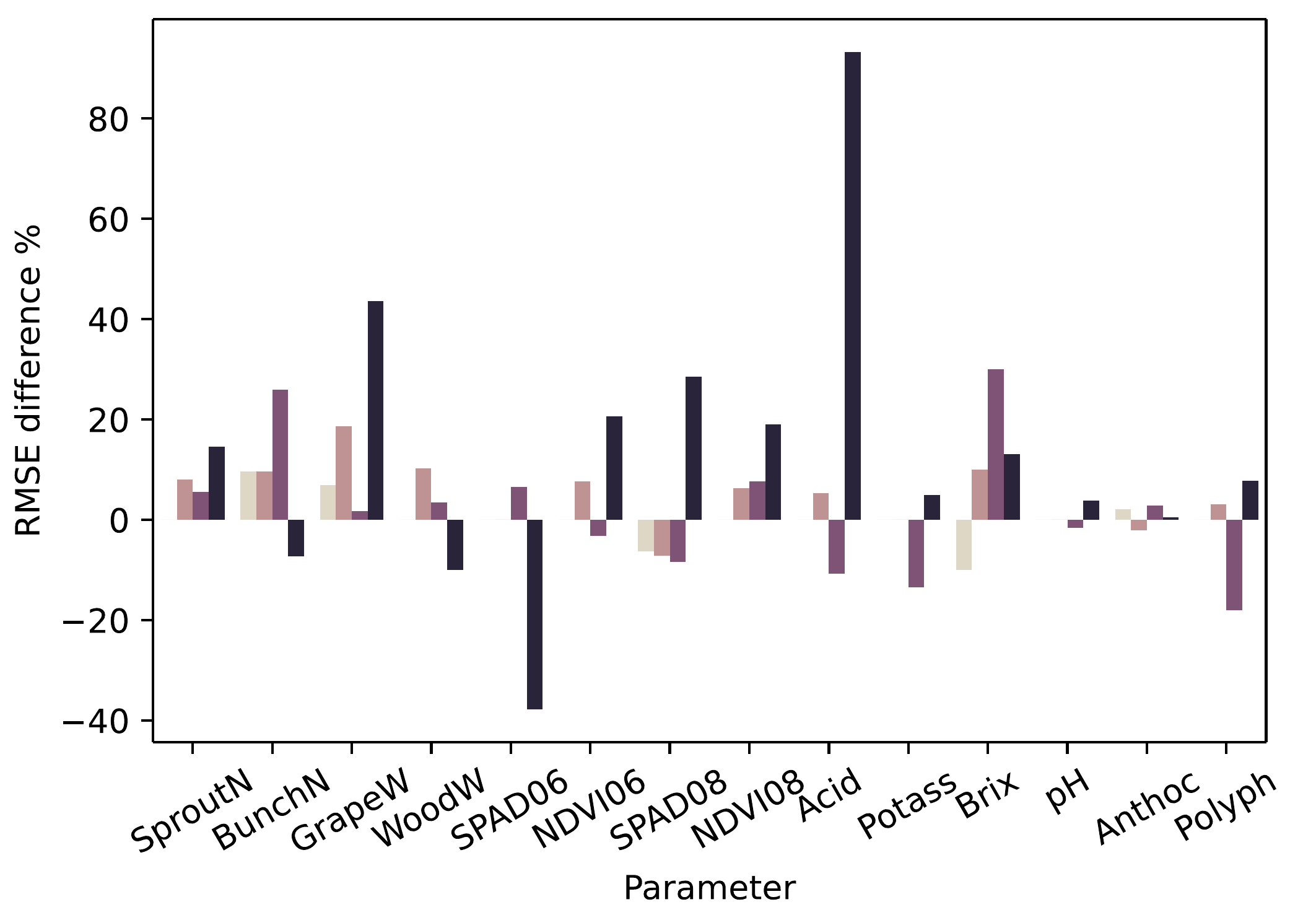}}
    \caption{The difference in RMSE percentage for two learning algorithms - Hill-Climbing and evolutionary algorithm.}
    \label{fig:two_algs}
\end{figure}

Then we compared the accuracy of the Hill-Climbing approach and the evolutionary algorithm. \cref{fig:two_algs} shows the difference in error in per cent for the Hill-Climbing algorithm and the evolutionary algorithm; if the difference is positive, then the Hill-Climbing algorithm error is less by the specified number of per cent, if the difference is negative, the Hill-Climbing error is greater by the specified number of per cent. From the results, we can't unequivocally say which algorithm is better. Hill-Climbing wins on some parameters, the evolutionary algorithm wins on others, but it is clear that the results of both algorithms are comparable. This diversity confirms the need to include both the greedy algorithm and the evolutionary algorithm in our approach to learning (\cref{fig:pipeline}).

\subsubsection{Measuring the learning time of algorithms}
Each algorithm considered in this paper can be composed of three blocks. The first element is the HC or evolutionary algorithm, the second is the mixed (M) or discrete (D) sore function, and the third is the mixed (M) or discrete (D) approach to parameter learning. Table \ref{table_time} shows the running times of the algorithm for each such possible combination on the same PC (i5-9300H CPU, 8GB RAM).

For Hill-Climbing in Table \ref{table_time}, one can notice a difference in the learning time of the structure depending on whether discrete or mixed parameter learning follows. This is explained by the smaller size of the DAGs space suitable for the mixed network.  Thus, mixed networks give results both better in quality and faster. Due to the greater randomness of the process for evolutionary algorithms, it isn't easy to draw any conclusion.

\begin{table}[!ht]
\caption{Structure learning time in seconds for both approaches, depending on the type of score function and dataset. For the evolutionary algorithm, the average time over 10 runs is taken. In the table header, the first D / M means learning the structure on discretized or mixed data and the second D / M means learning parameters on discretized or mixed data.}
\label{table_time}
\begin{center}
\begin{tabular}{|c|c|c|c|c|c|c|c|c|}
\hline
\multirow{2}{*}{Dataset}                                   & \multicolumn{4}{c|}{Hill-Climbing}                             & \multicolumn{4}{c|}{EVO}                                      \\ \cline{2-9} 

                                                             & D+D           & M+D           & D+M            & M+M           & D+D           & M+D           & D+M           & M+M           \\

\hline

HEALTHCARE & 1.633&	13.383&	0.861&	6.977&	83.844&	589.978&	96.094&	563.552 \\ \hline
SANGIOVESE & 11.593&	38.025&	9.841&	20.730&	132.936&	155.385&	132.305&	134.423 \\ \hline
MEHRA & 29.533&	111.275&	20.771&	62.300&	181.408&	864.444&	175.671&	928.631 \\ \hline
Reservoirs & 1.414&	33.994&	0.826&	11.137&	52.509&	344.306&	61.804&	302.141 \\ \hline
Social & 2.484&	78.676&	2.005&	53.106&	102.184&	866.546&	115.098&	795.717 \\

\hline
\end{tabular}
\end {center}
\end{table}

\subsubsection{Comparison of distributions}
\begin{figure}[!ht]
\centering
    \subfloat[]{\label{per_hc}\includegraphics[width=0.33\textwidth]{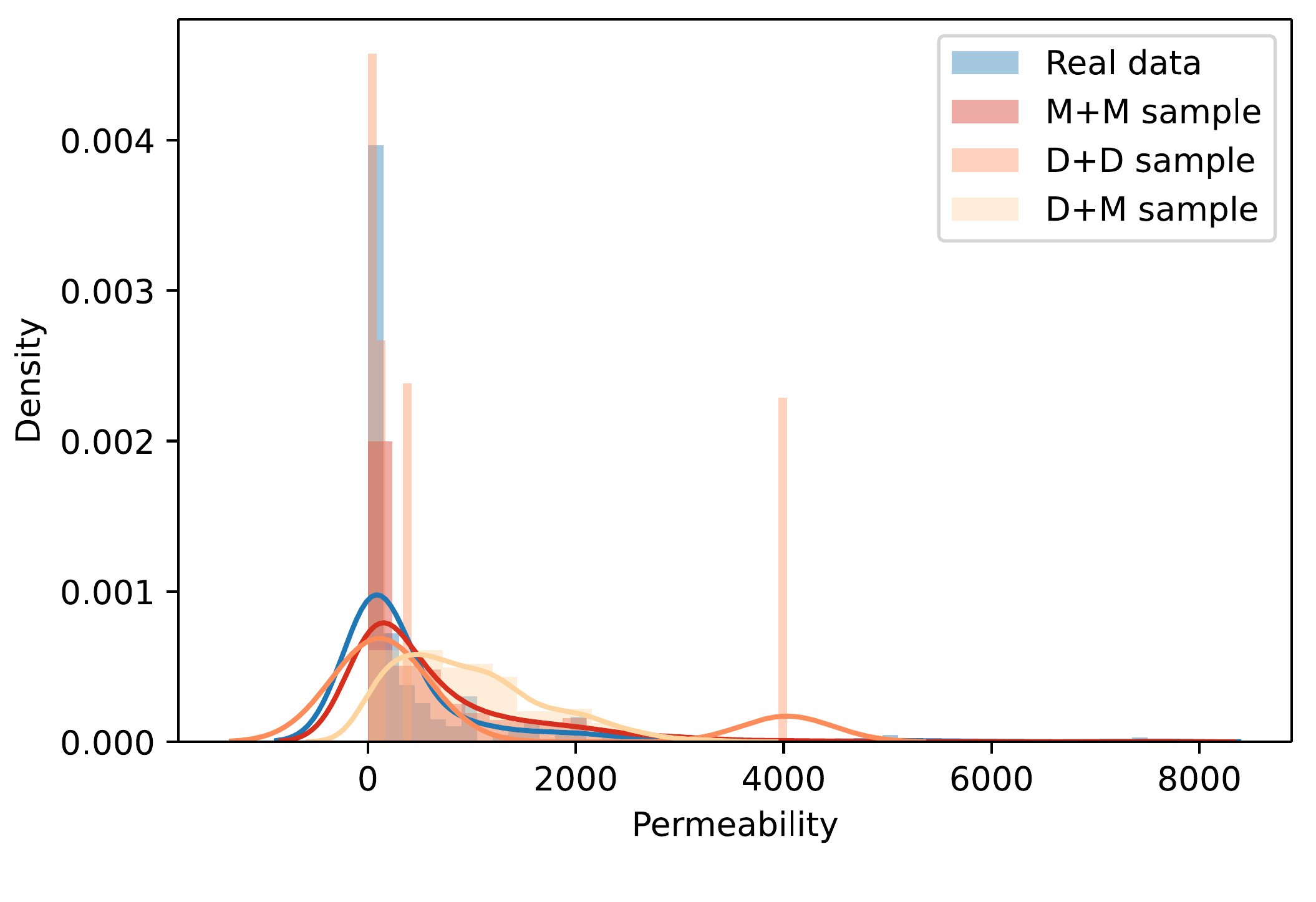}} 
    \subfloat[]{\label{depth_hc}\includegraphics[width=0.33\textwidth]{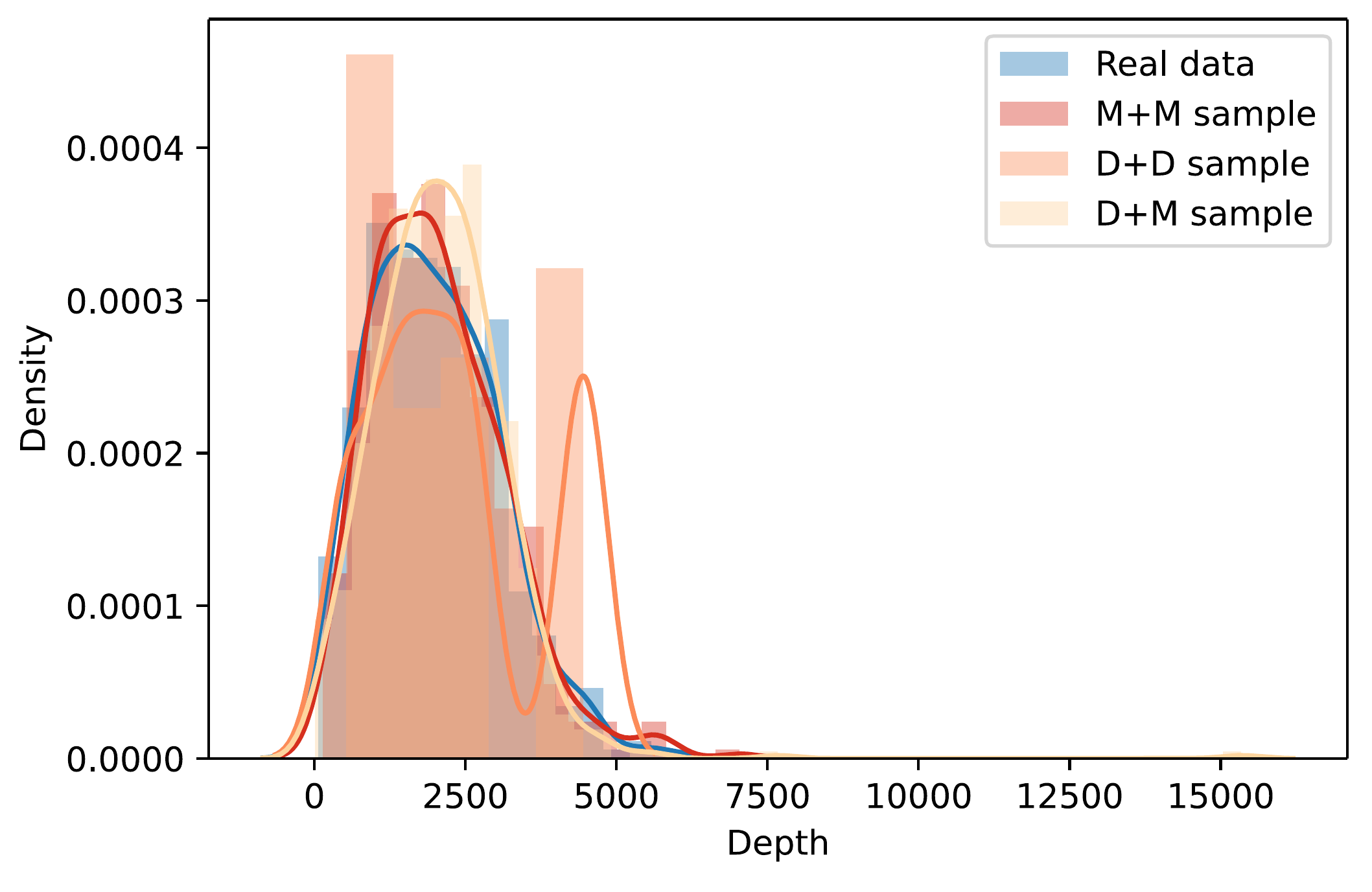}}
    \subfloat[]{\label{netpay_hc}\includegraphics[width=0.33\textwidth]{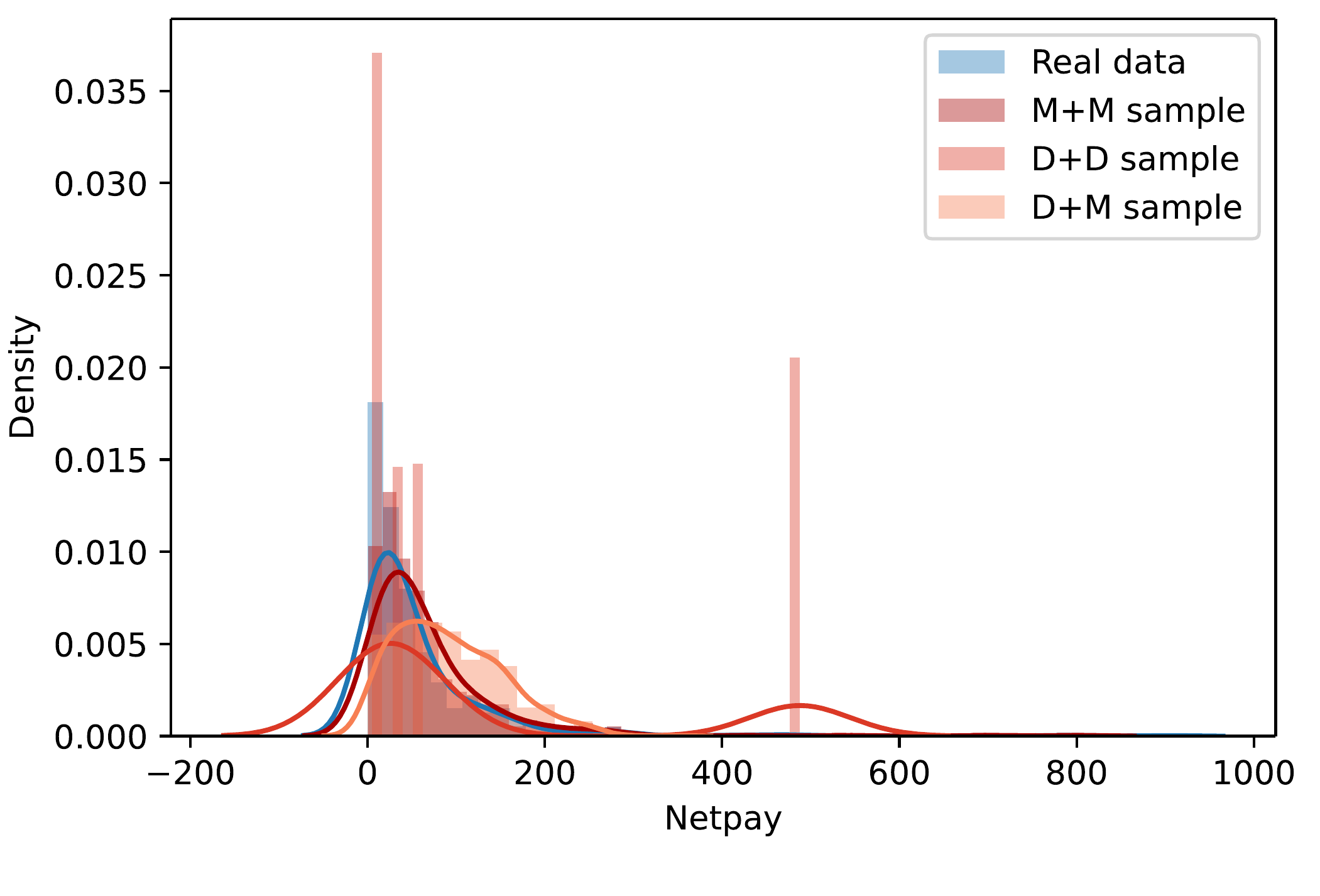}}\\
    \subfloat[]{\label{per_evo}\includegraphics[width=0.33\textwidth]{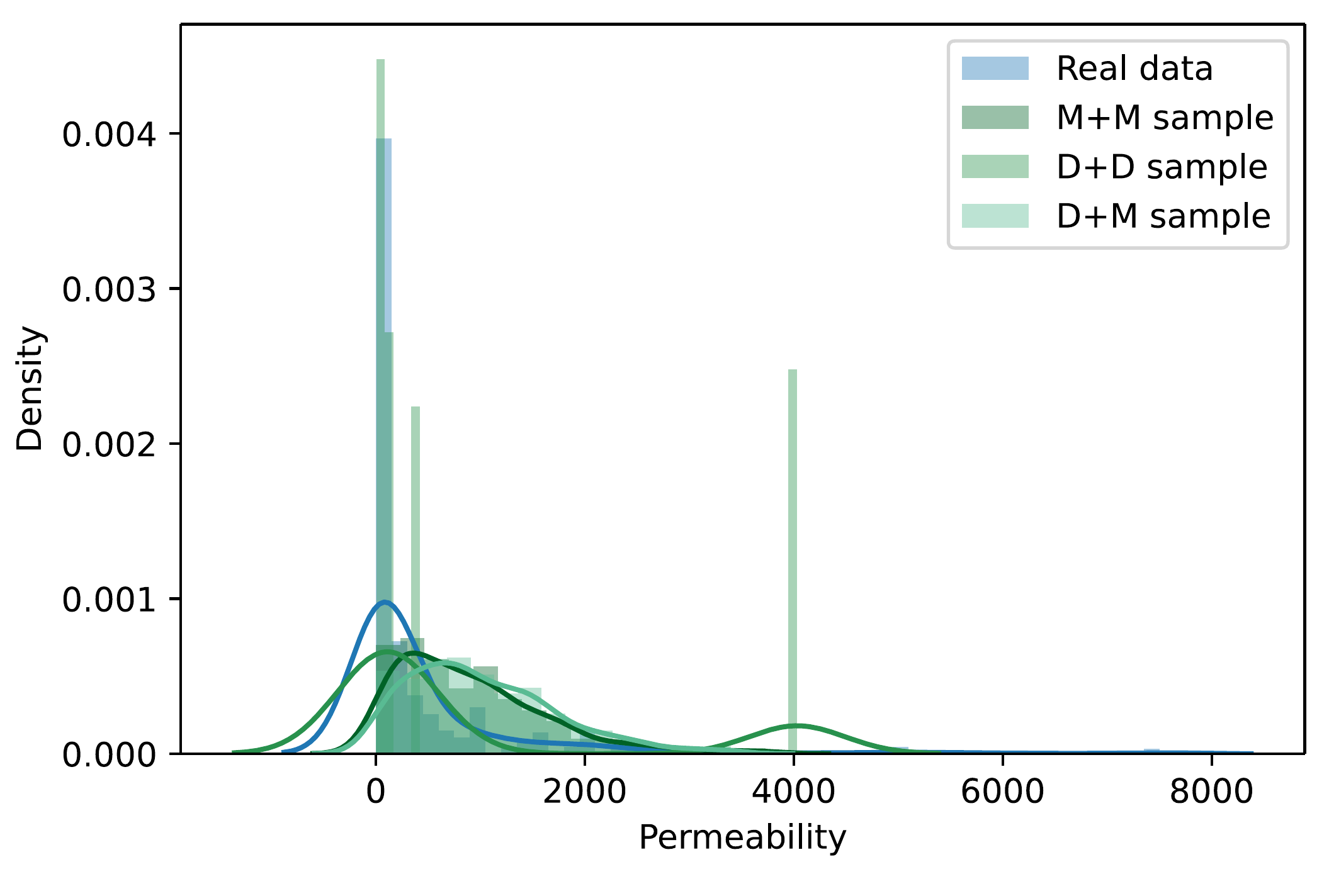}} 
    \subfloat[]{\label{depth_evo}\includegraphics[width=0.33\textwidth]{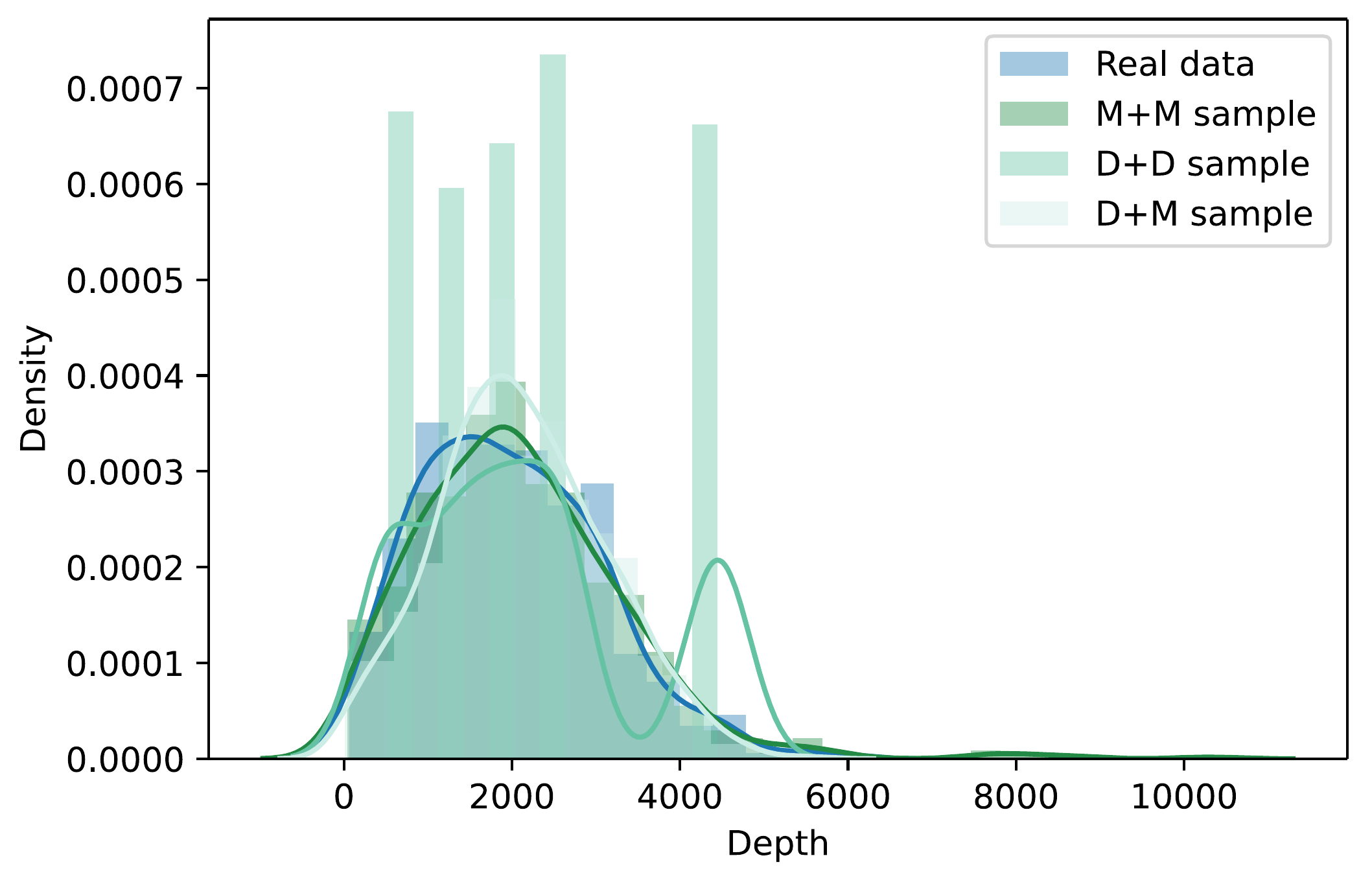}}
    \subfloat[]{\label{netpay_evo}\includegraphics[width=0.33\textwidth]{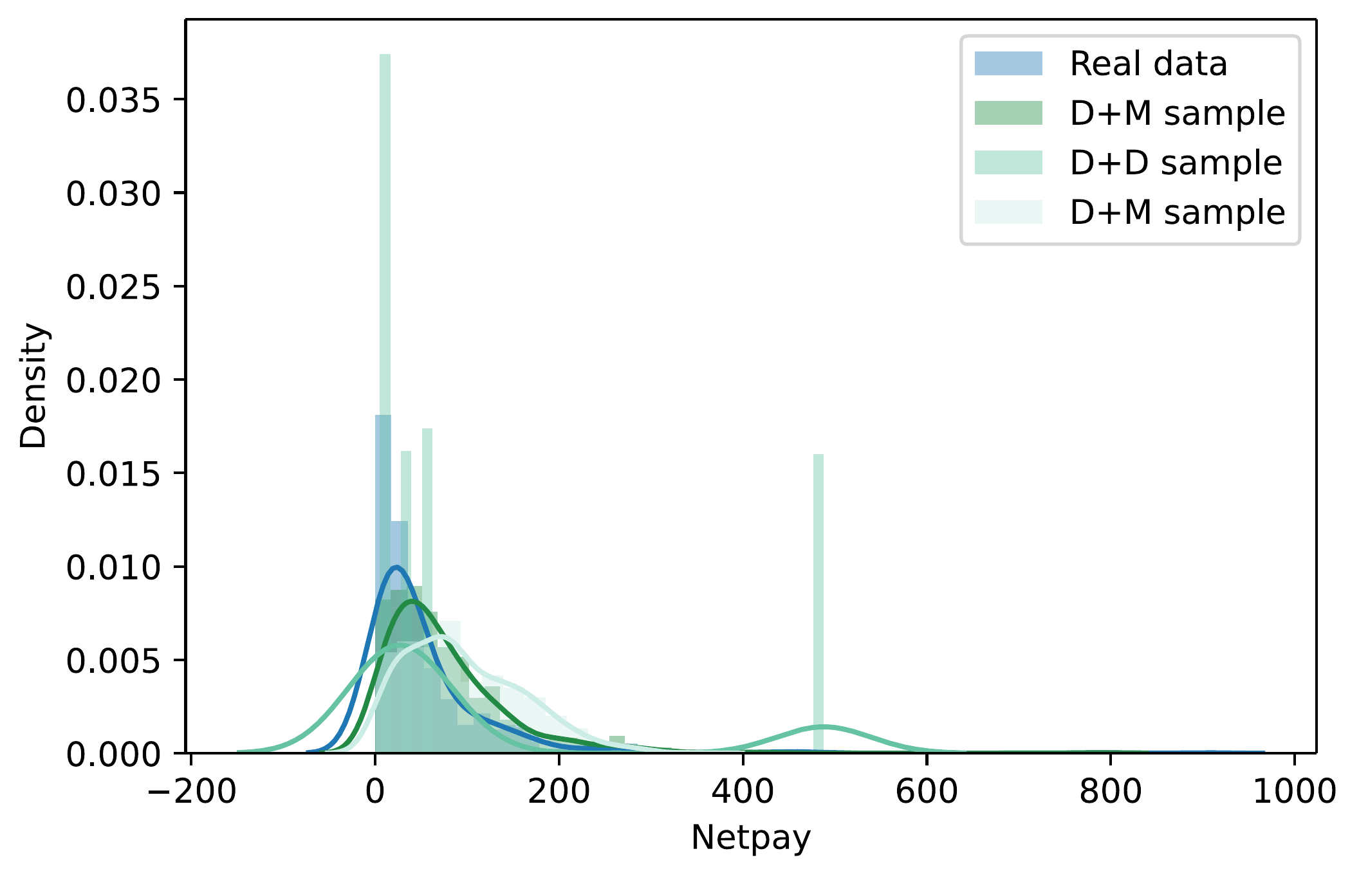}}
    \caption{Comparison of real distributions and sampled from Bayesian networks with different approaches to learning. The upper figures are the Hill-Climbing algorithm, the lower figures are the evolutionary algorithm.}
    \label{fig:dist}
\end{figure}

Firstly, to assess the quality of the resulting distributions when modelling with Bayesian networks, comparisons were made between real distributions and those sampled from Bayesian networks. The comparison was made on the example of some continuous nodes from real datasets (geo), since real data are the most difficult to model. \cref{fig:dist} shows a comparison of distributions for some parameters. It can be seen that the samples obtained from Bayesian networks by the M + M method are closest to the real distribution.

\section{Conclusion}
This paper presented the library MIxBN for learning Bayesian networks from mixed data that contains structural learning algorithm with mixed mutual information and mixed parameters learning with Gaussian approximation. Experiments have confirmed the effectiveness of the proposed algorithm developed within the library and have demonstrated an increase in the accuracy of modelling both on synthetic and real data. We also compared two algorithms for enumerating structures - Hill-Climbing and an evolutionary algorithm; it was found that each algorithm gives its gain on certain data, which confirms the need to add them to the library and increases the variety of proposed algorithms for learning Bayesian networks.

In the future, it would be interesting to expand the pool of investigated functions with log-likelihood, BIC, AIC based on the current modification of MI and compare the efficiency between themselves and for similar implementations. It is also planned to speed up and improve the quality of the evolutionary algorithm by using the starting structure obtained, for example, with HC. Another area for development would be to integrate block structure learning into our library, as this could reduce the number of operations on mixed data sets. We expect that this would lead to an increase in the speed of the algorithm. In the area of parameters learning and modelling, we plan to improve learning by mixtures of Gaussian distributions and by using polynomial regression instead of linear regression.

All source code and materials used in the paper are available in the repository \citep{github-sources}.

% \section{Acknowledgement}
% This research is financially supported by The Russian Scientific Foundation, Agreement \#....

%% References

%The citation must be used in following style: \cite{article-minimal} \cite{article-full} \cite{article-crossref} \cite{whole-journal}.

\bibliography{mybibliography}
\bibliographystyle{unsrtnat}




\clearpage

%%%% This page is for instructions only, once the article is finalize please omit the below text before creating the final PDF

\end{document}